\documentclass{article}

\usepackage[preprint]{neurips_2025}

\usepackage[utf8]{inputenc} 
\usepackage[T1]{fontenc}    
\usepackage{hyperref}       
\usepackage{url}            
\usepackage{booktabs}       
\usepackage{amsfonts}       
\usepackage{nicefrac}       
\usepackage{microtype}      
\usepackage{xcolor}         
\usepackage{graphicx}       
\usepackage{amsmath}
\usepackage{wrapfig}
\usepackage{appendix}
\usepackage{booktabs}
\usepackage{multirow}
\usepackage{epsfig}
\usepackage{colortbl}
\usepackage{xcolor}
\usepackage{makecell}
\usepackage{algorithm}          
\usepackage{algpseudocode}     
\usepackage{amsmath}           
\usepackage{amssymb} 

\usepackage{enumitem}
\usepackage{soul} 

\usepackage{color, xcolor}

\title{GRPO-CARE: Consistency-Aware Reinforcement Learning for Multimodal Reasoning}

\author{
\textbf{Yi Chen}$^{1,2}$, \textbf{Yuying Ge}$^{2}$\thanks{Corresponding Authors.}, \textbf{Rui Wang}$^{3}$, \textbf{Yixiao Ge}$^{2*}$, \textbf{Junhao Cheng}$^{2}$, \textbf{Ying Shan}$^2$, \textbf{Xihui Liu}$^{1*}$\\
    $^1$The University of Hong Kong,
    $^2$ARC Lab, Tencent PCG, 
    $^3$The Chinese University of Hong Kong\\
\small
{\href{https://github.com/TencentARC/GRPO-CARE}{\texttt{https://github.com/TencentARC/GRPO-CARE}}}
}

\begin{document}

\maketitle

\vspace{-27pt}
\begin{figure}[!htbp]
    \centering
    \includegraphics[width=\textwidth]{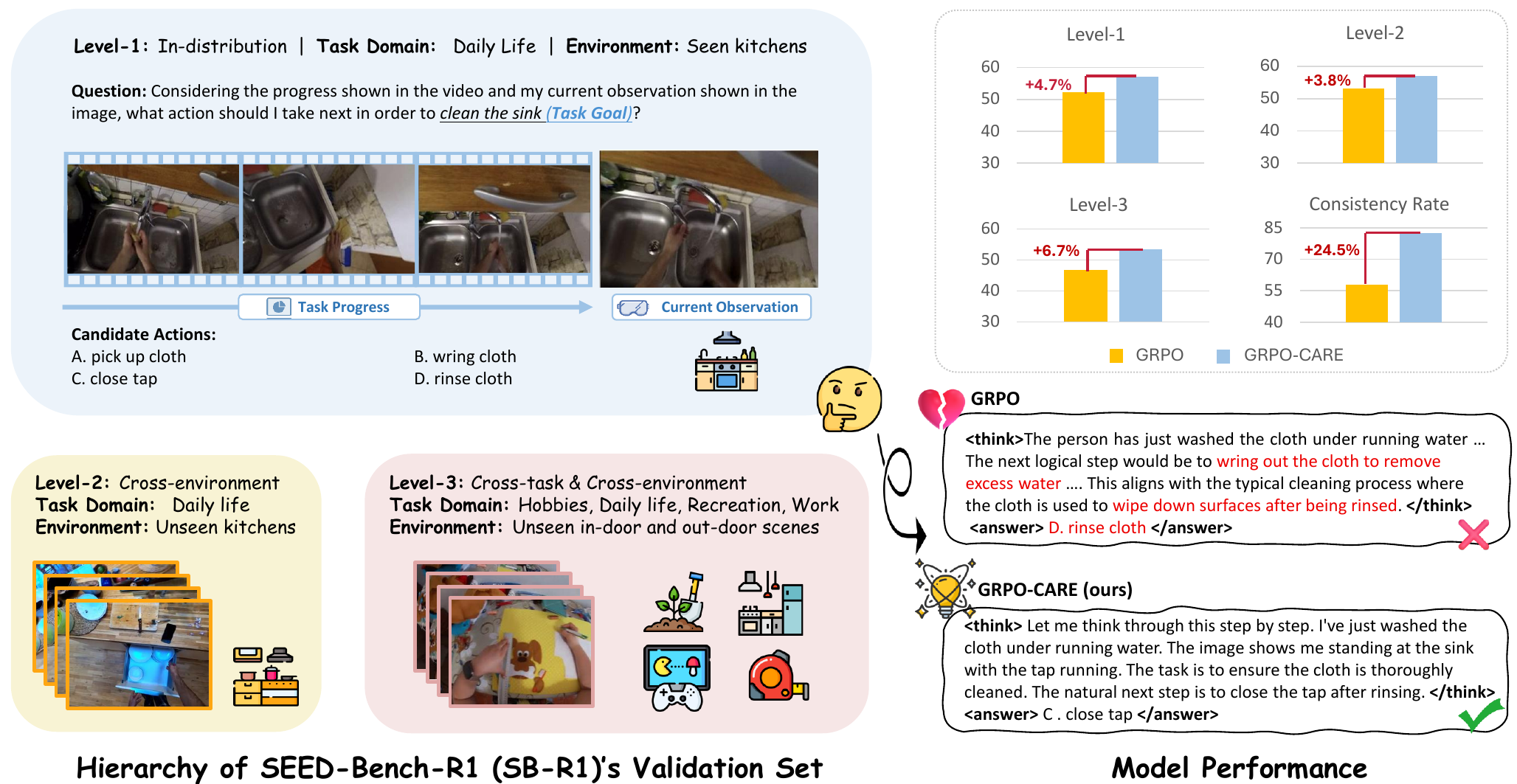}
    \vspace{-9pt}
    \caption{
    (a) \textbf{SEED-Bench-R1} (SB-R1) provides a systematic, three-level evaluation of post-training methods for MLLMs in video understanding, encompassing tasks that require both perception and reasoning to tackle complex real-world scenarios. (b) Our analysis identifies a key limitation of standard outcome-supervised GRPO: while it improves answer accuracy, it often compromises logical consistency between reasoning and answers. By introducing an adaptive, group-relative consistency bonus via reference-likelihood calibration, our \textbf{GRPO-CARE} achieves higher answer accuracy across all difficulty levels and improves interpretability, as reflected by increased consistency rates.}
    \label{fig:teaser}
\end{figure}

\begin{abstract}
Recent reinforcement learning (RL) approaches, such as outcome-supervised GRPO, have advanced Chain-of-Thought reasoning in large language models (LLMs), yet their adaptation to multimodal LLMs (MLLMs) remains unexplored. 
To address the lack of rigorous evaluation for MLLM post-training methods—especially on tasks requiring balanced perception and reasoning—we present \textbf{SEED-Bench-R1}, a benchmark featuring complex real-world videos that demand intricate visual understanding and commonsense planning. SEED-Bench-R1 uniquely provides a large-scale training set and evaluates generalization across three escalating challenges: in-distribution, cross-environment, and cross-environment-task scenarios.
Using SEED-Bench-R1, we identify a key limitation of standard outcome-supervised GRPO: while it improves answer accuracy, it often degrades the logical coherence between reasoning steps and final answers, achieving only a 57.9\% consistency rate. We attribute this to (1) reward signals focused solely on final answers, which encourage shortcut solutions at the expense of reasoning quality, and (2) strict KL divergence penalties, which overly constrain model exploration and hinder adaptive reasoning.
To overcome these issues, we propose \textbf{GRPO-CARE}, a novel consistency-aware RL framework that jointly optimizes for both answer correctness and reasoning coherence, without requiring explicit process supervision. GRPO-CARE introduces a two-tiered reward: (1) a base reward for answer correctness, and (2) an adaptive consistency bonus, computed by comparing the model’s reasoning-to-answer likelihood (via a slowly-evolving reference model) against group peers. This dual mechanism amplifies rewards for reasoning paths that are both correct and logically consistent.
By replacing the KL penalty with an adaptive, group-relative consistency bonus, GRPO-CARE consistently outperforms standard GRPO on SEED-Bench-R1, achieving a 6.7\% performance gain on the most challenging evaluation level and a 24.5\% improvement in consistency rate. Furthermore, GRPO-CARE demonstrates strong transferability, improving model performance across diverse video understanding benchmarks.
Our work contributes a systematically designed benchmark and a generalizable post-training framework, advancing the development of more interpretable and robust MLLMs.
\end{abstract}
\vspace{-5pt}
\section{Introduction}\label{ck:contribution}
\vspace{-2pt}
Recent advances in the reasoning capabilities of Large Language Models (LLMs)~\citep{guo2025deepseek, openai_o1, team2025kimi} have been largely driven by improvements in long Chain of Thought (CoT) generation. Among various enhancement strategies, reinforcement learning (RL)~\citep{shao2024deepseekmath, ouyang2022training, yeo2025demystifying} has emerged as a powerful post-training technique, enabling LLMs to refine their CoT reasoning through self-improvement guided by verifiable rewards. This leads to models that excel at solving complex problems and generalize well to out-of-distribution (OOD) tasks.
MLLMs extend LLMs by integrating modules for 
processing multimodal inputs, inheriting strong reasoning abilities while tackling richer, more complex data
~\citep{zhang2025r1,liu2025visual,meng2025mm}. However, existing evaluations for RL-like post-training methods for MLLMs tend to focus narrowly on either perception tasks (e.g., detection, grounding)~\citep{liu2025visual} or reasoning tasks (e.g., multimodal math problem solving)~\citep{huang2025vision}, or rely on broad general-domain training datasets without structured generalization assessment~\citep{video-r1}.

We argue that \textit{an ideal benchmark for post-training in multimodal understanding must balance perception and logical reasoning, while enabling rigorous evaluation of generalization}. Such a benchmark would foster models that integrate sophisticated perception and reasoning to achieve accurate, interpretable multimodal understanding, and robust performance in real-world scenarios.
To address this, we introduce \textbf{SEED-Bench-R1}, a challenging benchmark designed for systematic evaluation of post-training methods on video understanding. Built on prior benchmarks~\citep{chen2023egoplan,qiu2024egoplan} using realistic egocentric videos capturing everyday human activities~\citep{Damen2022RESCALING,grauman2022ego4d}, SEED-Bench-R1 requires models to comprehend open-form task goals, track long-horizon visual progress, perceive complex environmental cues, and reason about next actions using world knowledge, as shown in Fig.~\ref{fig:teaser}. Crucially, it features a three-level hierarchy for generalization assessment: Level-1 (in-distribution), Level-2 (OOD cross-environment), and Level-3 (OOD cross-environment-task), supported by large-scale training data and verifiable ground-truth answers suitable for RL.

Using SEED-Bench-R1, we conduct a comprehensive study comparing representative post-training methods. 
Our experiments confirm that RL—specifically GRPO based on outcome supervision~\citep{shao2024deepseekmath}—is highly data-efficient and significantly outperforms supervised fine-tuning (SFT) on both in-distribution and OOD questions.
However, we identify a key limitation: while outcome-supervised GRPO improves perception and answer accuracy for MLLMs, it often sacrifices logical coherence between reasoning chains and final answers, with a consistency rate of only 57.9\%.
This restricts interpretability and limits the potential performance ceiling. 
It originates from that optimizing solely for the final answer rewards creates a shortcut, 
where models prioritize answer correctness over maintaining logical coherence in reasoning steps. 
At the same time, strict KL divergence penalties excessively constrain the model's exploration, preventing adaptive adjustment of causal relationships between reasoning paths and answers, and further amplifying logical inconsistencies.

To overcome this, we propose \textbf{GRPO-CARE}, a novel RL framework with \textbf{C}onsistency-\textbf{A}ware \textbf{R}eward \textbf{E}nhancement that jointly optimizes answer correctness and logical consistency {without relying on explicit process supervision}. 
As illustrated in Fig.~\ref{fig:method}, in addition to the base reward for answer correctness, we introduce a consistency bonus derived from a slowly-updated reference model through \textit{likelihood calibration}. This bonus incentivizes the model to produce reasoning traces that are not only accurate but also logically coherent with the final answer.
Specifically, GRPO-CARE maintains a reference model by updating its parameters through an exponential moving average (EMA) of the online model’s parameters. This model calibrates reasoning-to-answer consistency likelihoods for high-accuracy samples generated by the online model. 
To evaluate alignment, we measure the likelihood that the reference model reproduces the same answer when given the reasoning trace and multimodal question inputs. 
Based on this likelihood, we assign a sparse bonus to samples that demonstrate both high accuracy and strong consistency within their group. 
By removing the KL divergence penalty and replacing it with an \textit{adaptive, group-relative consistency bonus}, GRPO-CARE encourages more effective exploration of coherent reasoning paths that lead to accurate answers.

Extensive evaluation on SEED-Bench-R1 demonstrates that GRPO-CARE consistently outperforms standard GRPO across all difficulty levels, especially in challenging OOD scenarios, improving performance by 6.7\% on the most difficult Level-3 evaluation and increasing the consistency rate by 24.5\%. Ablation studies confirm that the consistency-aware reward is critical for balancing overall performance and reasoning interpretability, surpassing alternative KL-based and reward-based baselines. Furthermore, GRPO-CARE-trained models show strong transferability to diverse general video understanding benchmarks, validating the robustness and generality of our approach.

In summary, our main contributions include:
\begin{itemize}[leftmargin=*,noitemsep,topsep=2pt]
\item We introduce {SEED-Bench-R1}, 
a novel benchmark that balances perception and reasoning, with rigorous hierarchical generalization evaluation for multimodal video understanding.

\item We conduct a systematic experimental analysis of post-training methods for MLLMs, revealing the limitations of current outcome-supervised GRPO in maintaining logical coherence.

\item We propose {GRPO-CARE}, a novel RL framework with a consistency-aware reward that significantly improves reasoning coherence and overall performance without explicit process supervision.
\end{itemize}

\vspace{-5pt}
\section{Related Work}
\vspace{-5pt}
\textbf{RL for LLMs/MLLMs.} 
RL from human feedback (RLHF) aligns LLM outputs with human preferences via reward models trained on human preference data~\citep{ouyang2022training,ppo}. To enhance complex reasoning, generating long CoT is effective~\citep{guo2025deepseek,team2025kimi,openai_o1}. RL methods like GRPO~\citep{shao2024deepseekmath} and its variants DAPO~\citep{dapo} and Dr.GRPO~\citep{liu2025understanding} optimize CoT generation using outcome-based rewards. However, outcome-only supervision can yield inconsistent reasoning despite correct answers. Addressing this, some works train additional process supervision reward models with costly step-wise annotations~\citep{prm,uesato2022solving,chenalphamath,luo2024improve,wang2023math}, incorporate LLM judges~\citep{gao2024llm,xia2025evaluating,zhang2024generative}, or adaptive regularization via EMA-updated reference models~\citep{warp}. In MLLMs, outcome-based RL may cause ``Thought Collapse,'' mitigated by stronger correctors~\citep{gtr} or step-wise reward matching~\citep{zhang2025r1}. Our GRPO-CARE employs a slowly updated reference model to provide bonus feedback on logically consistent and accurate responses, improving reasoning and accuracy without extra annotations or stronger correctors.

\textbf{Benchmarks for MLLM Post-training.} 
Recent RL-based post-training methods for MLLMs have primarily targeted image tasks—from perception (e.g., classification) to reasoning (e.g., visual math)~\citep{huang2025vision,liu2025visual,zhang2025r1,sun2024mm}. In contrast, video understanding, a more complex and general scenario, remains underexplored. 
Early RL-based efforts on video benchmarks~\citep{wang-2025-open-r1-video,zhao2025r1} are limited by narrow tasks (e.g., emotion recognition)~\citep{liu2022mafw,jiang2020dfew} or scarce training data~\citep{wu2024longvideobench}, hindering scalable analysis. Existing benchmarks~\citep{li2024mvbench,liu2024tempcompass,fang2024mmbench} mostly evaluate models post-trained on diverse general-domain data (e.g., Video-R1~\citep{video-r1}) but lack rigorous generalization assessment. 
To date, no comprehensive benchmark provides (1) large-scale training data for robust post-training, (2) structured validation sets across multiple generalization levels, and (3) multimodal questions balancing perception and reasoning in real-world scenarios. To address this, we propose SEED-Bench-R1, 
a video understanding benchmark with large-scale training data and a validation set partitioned into three generalization tiers, enabling comprehensive evaluation of MLLM post-training methods.

\vspace{-5pt}
\section{Pilot Study with SEED-Bench-R1}
\vspace{-5pt}
\subsection{SEED-Bench-R1}

\begin{table}[!t]
\caption{Data statistics of SEED-Bench-R1, which consists of a training set and a hierarchical three-level validation set for in-distribution, cross-environment, and cross-environment-task evaluations. }
\resizebox{\textwidth}{!}{%
\small
\centering
\begin{tabular}{ccccccc}
\toprule
\textbf{Split} & \textbf{\# Samples} & \textbf{Domain} & \textbf{Cross-Environment} & \textbf{Cross-Task} & \textbf{Video Source} & \textbf{Benchmark Source} \\ 
 \midrule
Train & 50,269 & Daily life & - & - & Epic-Kitchens & EgoPlan-Bench \\ 
Val-L1 & 2,432 & Daily life & $\times$ & $\times$ & Epic-Kitchens & EgoPlan-Bench \\ 
Val-L2 & 923 & Daily life & $\surd$ & $\times$ & Ego4D & EgoPlan-Bench \\ 
Val-L3 & 1,321 & \makecell{Hobbies, Daily life,\\ Recreation, Work} & $\surd$ & $\surd$ & Ego4D & EgoPlan-Bench2 \\
\bottomrule
\vspace{-22pt}
\end{tabular}}

\label{tab:benchmark_statistics}
\end{table}

\textbf{Benchmark Overview.} As shown in Fig.~\ref{fig:teaser}, SEED-Bench-R1 is a benchmark designed for systematically studying the impact of post-training methods for MLLMs on video understanding. 
Based on previous work, EgoPlan-Bench~\citep{chen2023egoplan} and EgoPlan-Bench2~\citep{qiu2024egoplan}, SEED-Bench-R1 features 1) intricate visual input from the real world, 2) diverse questions requiring logical inference with common sense to solve practical tasks, 3) rigorous partition of validation sets to assess the robustness and generalization abilities of MLLMs across different levels, and 4) large-scale automatically constructed training questions with easily verifiable ground truth answers.

\textbf{Visual Inputs and Question Design.} As shown in Fig.~\ref{fig:teaser}, the visual inputs and questions of SEED-Bench-R1 are grounded from realistic egocentric videos~\citep{Damen2022RESCALING,grauman2022ego4d} capturing everyday human activities. To answer questions in SEED-Bench-R1 correctly, the model must be capable of understanding open-form task goals, tracking long-horizon task progress, perceiving real-time environment state from an egocentric view, and utilizing inherent world knowledge to reason about the next action plan. The ground-truth answer comes from the actual next action occurring right after the current observation in the original uncropped video, with the negative options sampled from the same video. This challenging setting of candidate options demands a deep understanding of the environment state from dynamic visual input and world knowledge, such as action order dependency, rather than just the semantic meanings of task goals and actions, to discern the correct action plan. Moreover, the derivation of golden answers is traceable and easy to verify.

\textbf{Dataset Composition and Validation Levels.} As listed in Tab.~\ref{tab:benchmark_statistics}, we provide both training and validation datasets to benefit community research. The training dataset is automatically constructed using Epic-Kitchens~\citep{Damen2022RESCALING} videos recording daily life household tasks in kitchen environments. The validation dataset has undergone strict human verification to ensure correctness and is divided into three levels. The Level-1 (L1) questions are created using the same video source as the training data, representing in-distribution evaluation scenarios where the visual environments and task goals share overlaps with the training data. The Level-2 (L2) questions cover similar task goals as L1, but the visual observations are recorded in unseen kitchen environments by new participants from the Ego4D~\citep{grauman2022ego4d} team. 
The Level-3 (L3) validation subset utilizes the full set of Ego4D videos beyond the kitchen-specific subset. It 
contains general-domain questions spanning hobbies, recreation, and work, in addition to daily life. The visual inputs come from a variety of indoor and outdoor environments, posing greater challenges for testing the models' generalization abilities.

\vspace{-5pt}
\subsection{Experiment Setup}\label{ck:train-details-2}
We use Qwen2.5-VL-Instruct-7B~\citep{Qwen2.5-VL} as the backbone to study post-training methods in advancing the model performance on SEED-Bench-R1. 
We adopt outcome-supervised GRPO~\citep{shao2024deepseekmath} as a representative RL method and compare it with SFT. 
Both RL and SFT utilize 6k out of the 50k training samples from SEED-Bench-R1 for a pilot study.
To enhance training efficiency, we limit each video to 16 frames at resolution $128 \times 28 \times 28$, and append a frame indicating the current observation as an additional input.
For SFT, training data is augmented with CoT reasoning distilled from Qwen2.5-VL-Instruct-72B and 7B via rejection sampling. 
GRPO uses outcome supervision with rule-based rewards, eliminating the need for explicit CoT annotations. Following DeepSeek-R1~\citep{guo2025deepseek}, the model outputs reasoning within \texttt{<think> </think>} tags and the final answer within \texttt{<answer> </answer>} tags.

Given a multimodal question \( x \sim \mathcal{D} \), GRPO samples \( G \) responses \(\{o_g=(\tau_g, a_g)\}_{g=1}^G\) from the policy \( \pi_{\theta_{\text{old}}} \), where \(\tau_g\) and \(a_g\) are the reasoning process and the corresponding final answer, respectively. Unlike SFT, GRPO does not rely on predefined responses. The policy is optimized by maximizing:
\vspace{-1pt}
\setlength{\abovedisplayskip}{5pt}
\setlength{\belowdisplayskip}{5pt}
{\small
\begin{align*}
\mathcal{J}_\text{GRPO}(\theta) = \mathbb{E}_{x, \{o_g\}} \frac{1}{G} \sum_{g=1}^G \frac{1}{|o_g|} \sum_{i=1}^{|o_g|} \min \Bigg[ & \frac{\pi_{\theta}(o_{g,i} | x, o_{g,<i})}{\pi_{\theta_{\text{old}}}(o_{g,i} | x, o_{g,<i})} \hat{A}_{g,i}, \\
& \operatorname{clip}\left(\frac{\pi_{\theta}(o_{g,i} | x, o_{g,<i})}{\pi_{\theta_{\text{old}}}(o_{g,i} | x, o_{g,<i})}, 1-\varepsilon, 1+\varepsilon\right) \hat{A}_{g,i} \Bigg] - \beta \mathbb{D}_{KL}[\pi_{\theta} || \pi_{\text{ref}}]
\end{align*}
}

Here, \(\varepsilon\) and \(\beta\) are hyperparameters, and \(\mathbb{D}_{KL}\) is the KL divergence between the trained policy \(\pi_\theta\) and a reference policy \(\pi_{\text{ref}}\). The per-token advantage \(\hat{A}_{g,i}\) is set to the normalized reward \(\widetilde{r}_g\), computed from rule-based rewards \(r_g\) (e.g., \(r_g=1\) if the extracted answer matches ground truth, else 0) across the group $\hat{A}_{g,i} = \widetilde{r}_g = \frac{r_g - \operatorname{mean}(\{r_1, ..., r_G\})}{\operatorname{std}(\{r_1, ..., r_G\})}$.

\vspace{-5pt}
\subsection{Result Analysis}

\begin{table}[!t]
\small
\centering
\caption{Performance comparison on SEED-Bench-R1's hierarchical validation set. 
}
\resizebox{\textwidth}{!}{%
\begin{tabular}{lccccccc}
\toprule
\multirow{2}{*}{{Models}} & {{L1 (In-Distribution)}} & {{L2 (Cross-Env)}} & \multicolumn{5}{c}{{L3 (Cross-Task, Cross-Env)}} \\
\cmidrule(r){2-2} \cmidrule(r){3-3} \cmidrule(r){4-8}
 & {Daily Life} & {Daily Life} & {Daily Life} & {Hobbies} & {Recreation} & {Work} & {Overall} \\
\midrule
Qwen2.5-VL-7B & 38.4 & 40.1 & 35.8 & 31.2 & 26.8 & 28.5 & 31.3 \\
SFT & 46.2 & 46.3 & 46.7 & 41.7 & 44.3 & 38.4 & 42.7 \\
GRPO & 52.3 & 53.2 & 51.9 & 43.7 & 55.2 & 39.4 & 46.7  \\
GRPO-CARE (ours) & \textbf{57.0} & \textbf{57.0} & \textbf{57.6} & \textbf{51.2} & \textbf{57.4} & \textbf{48.5} & \textbf{53.4} \\
\bottomrule
\end{tabular}}
\vspace{-5pt}
\label{tab:seedbench_r1_performance}
\end{table}

Tab.~\ref{tab:seedbench_r1_performance} summarizes the performance of MLLMs trained with various methods on SEED-Bench-R1. Notably, compared to SFT, reinforcement learning with GRPO significantly improves data efficiency and boosts MLLM performance on both in-distribution (L1) and OOD (L2, L3) questions, despite relying only on a simple outcome-based reward without specialized CoT annotations.

\begin{figure}[!t]
    \centering
    \includegraphics[width=0.97\textwidth]{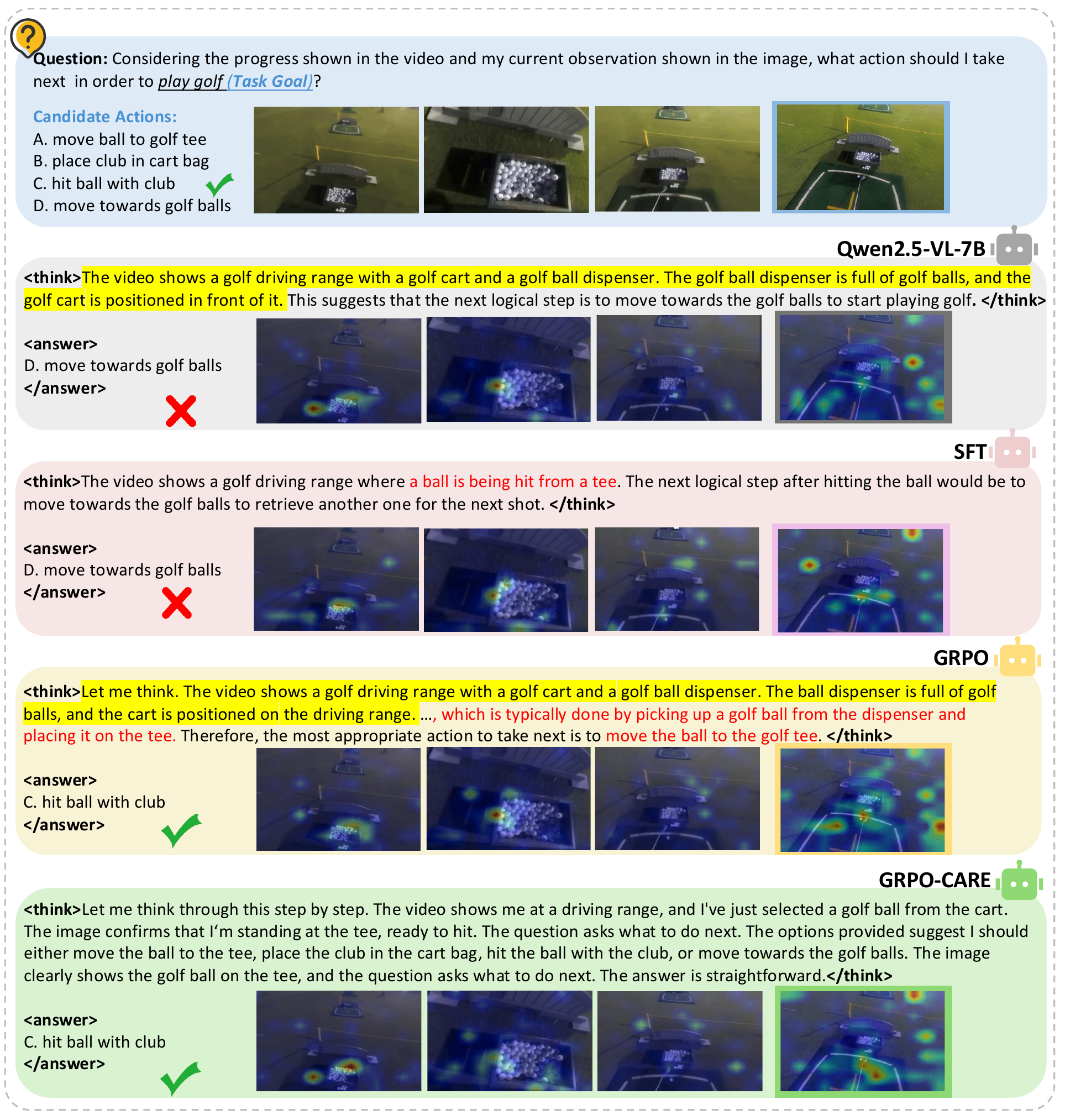}
    \caption{
    Case study of an L3 question from SEED-Bench-R1, showing a video of task progress, a final observation image, and attention maps (output-to-visual tokens).
    The \sethlcolor{pink}\hl{SFT} model tends to memorize reasoning patterns and exhibits perceptual hallucinations.
    The \sethlcolor{yellow}\hl{GRPO} model attends more comprehensively to the highlighted key visual observation while lacking logical consistency in the generated content. 
    The \sethlcolor{green}\hl{GRPO-CARE} model further balances visual perception and logical reasoning.
    }
    \label{fig:case_study_L3_1}
    \vspace{-10pt}
    
\end{figure}

Our analysis shows that GRPO mainly enhances perceptual abilities rather than reasoning. As shown in Fig.~\ref{fig:case_study_L3_1}, the SFT-trained model is more prone to perceptual hallucinations, such as describing ``a ball being hit from a tee'' when this event does not occur. Attention map analysis reveals that GRPO-trained models generate CoT tokens that act as dynamic queries, attending to visual content more thoroughly—especially in OOD scenarios. For example, the GRPO model better highlights key visual observations and allocates more attention to critical objects (e.g., the ball on the tee), even if these are not explicitly referenced in the reasoning. We hypothesize that RL methods like GRPO encourage broader visual exploration via CoT, while SFT tends to produce superficial, pattern-memorized CoT with limited visual grounding. This likely underpins GRPO’s superior generalization.

However, outcome-supervised GRPO training for MLLMs has key limitations: unlike LLMs, MLLM reasoning does not improve proportionally during RL, often resulting in logical inconsistencies. While the GRPO-trained model frequently reaches correct answers, its CoT reasoning often lacks coherence. For instance, as shown in Fig.~\ref{fig:case_study_L3_1}, initial reasoning steps mirror those of the base model (Qwen2.5-VL-7B), but later steps diverge and may contradict each other—e.g., suggesting ``move the ball to the golf tee'' but ultimately answering ``hit ball with club.'' Such inconsistencies, though sometimes yielding correct answers, undermine transparency.

Limited reasoning also constrains overall performance, as effective reasoning is crucial for integrating world knowledge with perception. For example, in Fig.~\ref{fig:teaser}, the GRPO model correctly identifies ``running water'' but fails to infer that the next logical step after cleaning is ``turning off the faucet.'' These reasoning-answer mismatches further complicate interpretability.

\section{Consistency-Aware Reward-Enhanced GRPO for MLLMs (GRPO-CARE)} \label{ck:training-details}
While outcome-supervised GRPO enhances visual perception in MLLMs, our analysis on SEED-Bench-R1 uncovers a critical trade-off: it often produces less logically coherent reasoning chains, thereby limiting interpretability and performance. This issue arises from two main limitations. First, the standard reward focuses exclusively on final-answer accuracy, overlooking the quality of intermediate reasoning steps. This can incentivize shortcut solutions—correct answers reached via inconsistent reasoning. Second, the KL penalty disproportionately constrains reasoning traces, typically longer than answers, thereby stifling exploration of diverse and coherent reasoning paths.

\begin{figure}[!t]
    \centering
    \includegraphics[width=\textwidth]{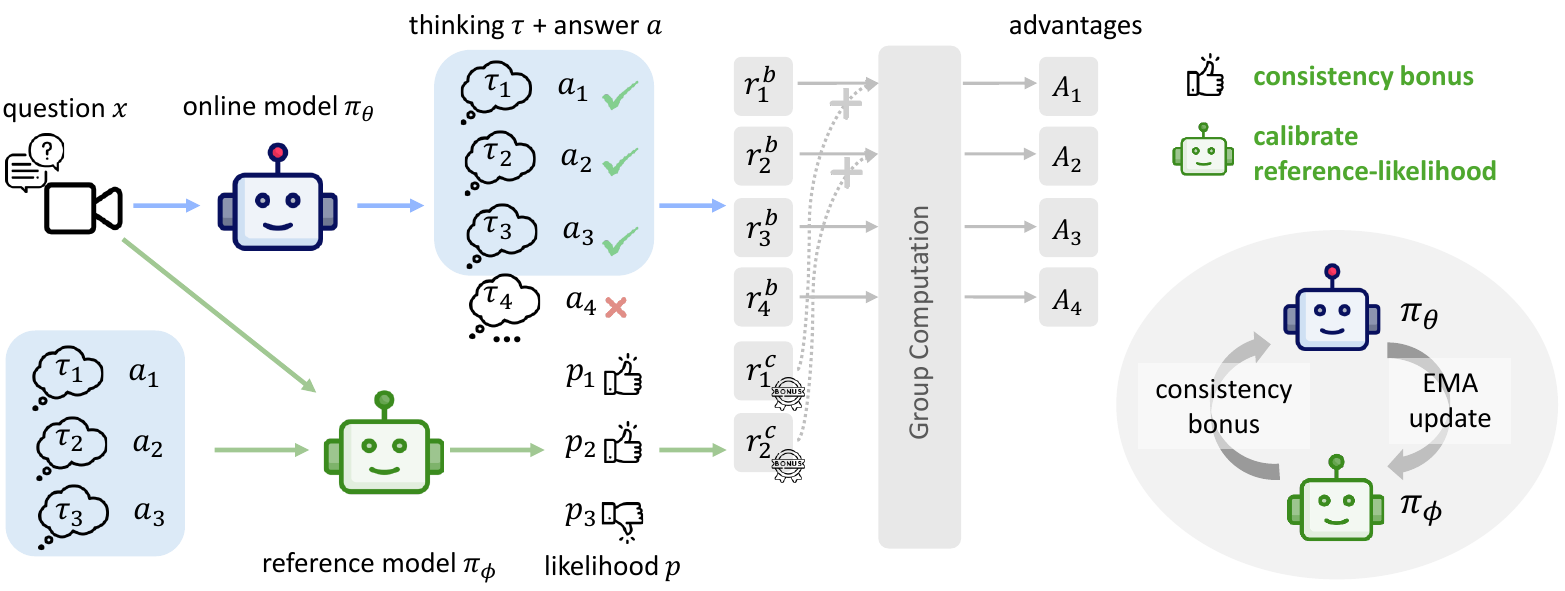}
    \caption{GRPO-CARE uses a two-tier reward system: a base reward for answer correctness ($r^b_*$) and an adaptive consistency bonus ($r^c_*$). The consistency bonus is given to high-accuracy samples whose reasoning-to-answer likelihood—estimated by a slowly updated (EMA) reference model—is higher than that of their group peers, conditioned on the multimodal question. The total reward, the sum of base and consistency rewards, is then used to compute advantages for updating the online model.
    }
    \label{fig:method}
    \vspace{-3mm}
\end{figure}

To address these challenges, we propose \textbf{GRPO-CARE} (\textbf{C}onsistency-\textbf{A}ware \textbf{R}eward \textbf{E}nhancement), a method that jointly optimizes for both answer correctness and logical consistency, without requiring explicit supervision of the reasoning process. As shown in Fig.~\ref{fig:method}, GRPO-CARE introduces a two-tiered reward system: a base reward for answer correctness, and an adaptive consistency bonus. The consistency bonus is calculated by comparing the likelihood that a reasoning trace leads to the correct answer, as estimated by a slowly evolving reference model. For each high-accuracy sample generated by the online model, this likelihood is compared with those of its peers within the same group, encouraging the exploration of reasoning traces that are logically consistent with correct answers.

The training process, detailed in Algorithm~\ref{algorithm1}, involves \textbf{two-stage filtering}. (1) First, we generate multiple reasoning traces per input and retain only those that exceed an accuracy baseline. (2) For these high-accuracy candidates, we assess how well each reasoning trace supports the final answer by calibrating its likelihood using a slowly evolving reference model.

\textbf{Reference Model and Likelihood Calibration.} The key insight is that \textit{a stable reference model—when conditioned on the online model’s reasoning trace—should assign a higher likelihood to the correct answer if the reasoning is logically grounded in the multimodal input}. Specifically, the reference model is initialized from the same pretrained weights as the online model and updated via exponential moving average (EMA) to ensure stable likelihood estimation and self-adaptation. To avoid reinforcing ``consistent-but-wrong'' reasoning, we compute this likelihood only for trajectories with correct answers. Additionally, we cap the likelihood at a maximum threshold to prevent over-optimization toward artificially high values. 

\textbf{Consistency Bonus Calculation.} Based on the clipped reference likelihoods, we compute a \textit{group-relative consistency baseline} as the mean clipped likelihood (minus a small margin to avoid penalizing near-average samples). Trajectories that exceed this baseline receive a sparse \textit{consistency bonus}, weighted by their accuracy, ensuring that rewards prioritize both correctness and logical coherence.

\textbf{Model Update.} 
To promote exploration of diverse reasoning paths, we remove the KL penalty from the GRPO training objective. Instead, we rely on the consistency bonus—added to the base reward to form the total reward—to guide online model updates toward higher-quality outputs.
The reference model is updated via EMA every few steps, allowing it to gradually inherit improvements from the online model (e.g., better visual grounding or more complex reasoning) while maintaining stability against sampling noise. This balanced optimization process enhances multimodal understanding without sacrificing logical consistency, ultimately improving both performance and interpretability.

\begin{algorithm}[!htbp]
\small
\caption{Consistency-Aware Reward Enhanced GRPO}
\begin{algorithmic}[1]
\Require
\Statex $\pi_{\theta}$: Online policy model (initialized from pretrained weights)
\Statex $\pi_{\phi}$: Reference model with EMA updates ($\phi \leftarrow \theta$ initially)
\Statex $\mathcal{D}$: Multimodal training dataset $\{(x,y^*)\}$ 
\Statex $\lambda_{\text{cons}}$: Consistency reward coefficient (e.g., 0.5)
\Statex $\gamma_{\text{acc}}$: Minimum accuracy threshold (e.g. 0.1)
\Statex $\gamma_p$: Maximum likelihood threshold (e.g. 0.95)
\Statex $\epsilon_p$: Consistency margin (e.g. 0.01)

\Procedure{Training}{$\pi_{\theta}, \mathcal{D}, T$}
\For{$t \gets 1$ to $T$}
    \For{each multimodal input $x$ in batch $\mathcal{D}$}
        \State \textbf{Phase 1: Trajectory Generation \& Reward Computation}
        \State Generate $G$ reasoning traces + answers: $\{\tau_g, a_g\}_{g=1}^G \sim \pi_{\theta}(\cdot|x)$
        \State Compute accuracy rewards: $r_{\text{acc},g} = \text{accuracy\_score}(a_g, y^*)$
        \State Compute format rewards: $r_{\text{fmt},g} = \text{format\_score}(\tau_g, a_g)$
        
        \State \textbf{Phase 2: Relative High-Accuracy Trajectory Selection}
        \State Calculate relative accuracy baseline: $\hat{r}_{\text{acc}} = \max(\mathbb{E}_g[r_{\text{acc},g}], \gamma_{\text{acc}})$
        \State Select trajectories where $r_{\text{acc},g} \geq \hat{r}_{\text{acc}}$
        
        \State \textbf{Phase 3: Relative Consistency Evaluation}
        \For{selected trajectories $(\tau_g,a_g)$}
            \State Compute reference likelihood:
            $p_g = \frac{1}{|a_g|} \sum_{i=1}^{|a_g|} \pi_{\phi}(a_{g,i} \mid x, \tau_g, a_{g,<i})$

            \State Clip likelihood: $\tilde{p}_g = \min(p_g, \gamma_p)$
        \EndFor
        \State Calculate relative consistency baseline: $\hat{\mu}_p = \mathbb{E}_g[\tilde{p}_g] - \epsilon_p$
        \State Select consistent trajectories where $\tilde{p}_g \geq \hat{\mu}_p$
        
        \State \textbf{Phase 4: Enhanced Reward Calculation}
        \For{each trajectory $g$}
            \State $R_g = \underbrace{r_{\text{acc},g} + r_{\text{fmt},g}}_{\text{base reward}} + \underbrace{\lambda_{\text{cons}} \cdot r_{\text{acc},g} \cdot \mathbb{I}[\text{consistent}]}_{\text{consistency bonus}}$
            \State Normalize advantage: $\hat{A}_g = (R_g - \mu_R)/\sigma_R$
        \EndFor
    \EndFor
    
    \State \textbf{Phase 5: Model Update}
    \State Update $\pi_{\theta}$ via GRPO policy gradient (without KL penalty)
    
    \If{$t \bmod k = 0$} \Comment{EMA update every $k=10$ steps}
        \State $\phi \gets \alpha \phi + (1-\alpha)\theta$ \Comment{$\alpha$=0.995 typical}
    \EndIf
\EndFor
\State \Return optimized policy $\pi_{\theta}$
\EndProcedure
\end{algorithmic}
\end{algorithm}\label{algorithm1}

\subsection{Evaluation on SEED-Bench-R1}\label{ck:llm-eval}
We first evaluate our method on SEED-Bench-R1. As shown in Tab.~\ref{tab:seedbench_r1_performance}, GRPO-CARE significantly outperforms GRPO across all three difficulty levels, with a particularly notable improvement of nearly 10\% on the most challenging L3 evaluation in domains such as \textit{Hobbies} and \textit{Work}, which exhibit substantial distributional divergence from the training data.  

To thoroughly assess the effectiveness of GRPO-CARE, we compare it against two families of baseline methods: \textbf{KL-oriented baselines}, which modify the application of divergence constraints, and \textbf{reward-based alternatives}, which replace KL penalties with consistency-aware rewards.  

\begin{itemize}[leftmargin=*,noitemsep,topsep=2pt]
\item \textbf{KL-Oriented Baselines.} \textbf{1) KL-EMA} introduces an EMA-updated reference model for adaptive constraints. \textbf{2) KL-EMA-HA} selectively applies KL penalty only to high-accuracy samples, applying regularization on where alignment matters most. \textbf{3) SepKL-EMA-HA} further decomposes KL into separate terms for reasoning and answer tokens to alleviate disproportionately penalizing lengthy reasoning tokens while potentially overlooking answer-reasoning inconsistencies. \textbf{4) NoKL} removes the KL penalty, demonstrating the raw optimization potential absent any regularization.

\item \textbf{Reward-Based Alternatives.} \textbf{5) DenseCons} applies continuous likelihood weighting to derive dense consistency rewards: $r_{\text{cons}} = \lambda_{\text{cons}} \cdot r_{\text{acc}} \cdot p_\phi (a | x, \tau)$. \textbf{6) RefGen} takes a more explicit approach by having the reference model regenerate answers from sampled reasoning paths, using the regenerated answer's accuracy as the consistency signal: $r_{\text{cons}} = \lambda_{\text{cons}} \cdot \text{accuracy}({a' \sim \pi_\phi(\cdot|x,\tau)}, y^*)$. 
\end{itemize}

As shown in Tab.~\ref{tab:ablation_study}, we report both benchmark performance and the \textbf{consistency rate} between generated reasoning and final answers, where consistency is evaluated by GPT-4.1 to assess whether the reasoning sufficiently supports the answer. 
Our analysis shows that while the EMA-updated reference model improves both accuracy and consistency, restricting KL penalties to high-accuracy samples (\textit{KL-EMA-HA}) boosts in-domain (L1) results but slightly reduces OOD (L2/L3) generalization. Decomposing KL penalties (\textit{SepKL-EMA-HA}) mitigates reasoning-answer inconsistency, yielding minor gains on L2 but limited impact on L3. Notably, none of the KL-based variants outperform \textit{NoKL}, indicating that standard KL regularization may hinder the performance ceiling in this context.

\begin{wraptable}{r}{8cm}
\centering
\small
\vspace{-12pt}
\caption{Ablation studies on SEED-Bench-R1.}
\vspace{1em}
\label{tab:model_performance}
\begin{tabular}{lcccc}
\toprule
Models & L1 & L2 & L3 & Consistency \\
\midrule
GRPO & 52.3 & 53.2 & 46.7 & 57.9 \\
\midrule
KL-EMA & 54.7 & 54.1 & 49.4 & 60.0 \\
KL-EMA-HA & 55.1 & 53.8 & 49.2 & 61.7 \\
SepKL-EMA-HA & 54.8 & 54.9 & 47.5 & 76.8 \\
noKL & 55.4 & 54.4 & 51.3 & 70.0 \\
\midrule
DenseCons & 56.6 & 55.5 & 50.6 & 80.3 \\
RefGen & 55.2 & 54.2 & 49.4 & \textbf{86.4} \\
CARE (ours) & \textbf{57.0} & \textbf{57.0} & \textbf{53.4} & 82.4 \\
\bottomrule
\end{tabular}
\vspace{-8pt}
\label{tab:ablation_study}
\end{wraptable}

Among reward-based methods, \textit{DenseCons} surpasses \textit{NoKL} on L1 and L2 with improved consistency, but slightly underperforms on L3, likely due to over-reliance on reference model calibration. \textit{RefGen} greatly increases consistency but introduces instability from sampling-based answer regeneration, ultimately reducing overall performance.

Our proposed \textbf{GRPO-CARE} uses sparse consistency rewards to achieve robust improvements across all levels. Its two-stage filtering—leveraging adaptive EMA-updated reference likelihoods to provide relative, sparse feedback for high-accuracy samples—effectively enhances logical consistency and answer accuracy. This demonstrates that group-relative sparse rewards deliver more reliable learning signals, avoiding overfitting to imperfect likelihoods (as in DenseCons) or sampling noise (as in RefGen).

\begin{table}[!t]
\centering
\caption{Performance of different models on general video understanding benchmarks}
\label{tab:model_performance}
\resizebox{\textwidth}{!}{%
\begin{tabular}{lccccccc}
\toprule
Models & Frames & VSI-Bench & VideoMMMU & MMVU & MVBench & TempCompass & VideoMME \\
\midrule
GPT-4o~\cite{gpt4o} & - &  34.0 & 61.2 & 75.4 & - & - & 71.9  \\
\midrule
LLaMA-VID~\cite{llamavid} & - &  - & - & - & 41.9 & 45.6 & -  \\
VideoLLaMA2~\cite{cheng2024videollama}  & - &  - & - & 44.8 & 54.6 & - & 47.9  \\
LongVA-7B~\cite{longva}  & - & 29.2 & 23.9 & - & - & 56.9 & 52.6 \\
VILA-1.5-8B~\cite{lin2024vila}  & - & 28.9 & 20.8 & - & - & 58.8 & - \\
Video-UTR-7B~\cite{video-utr}  & - &  - & - & - & 61.1 & 62.5 & 56.0  \\
LLaVA-OneVision-7B~\cite{li2024llava}  & - & 32.4 & 33.8 & 49.2 & 56.7 & - & 58.2 \\
Kangeroo-8B~\cite{liu2024kangaroo}  & - & - & - & 61.1 & 62.5 & 69.9 & 55.4 \\
Video-R1-7B~\cite{video-r1} & 32 & \textbf{35.8} & \textbf{52.3} & 63.8 & 63.9 & 73.2 & 59.3  \\
\midrule
Qwen2.5-VL-7B & 32 & 30.1 & 48.1 & 60.0 & 59.0 & 72.6 & 56.6 \\
CARE-7B (SB-R1) & 32 & 34.3 & 51.6 & \textbf{66.2} & 63.2 & \textbf{74.3} & 58.1 \\
CARE-7B (Video-R1) & 32 & \textbf{35.8} & 50.4 & {65.8} & \textbf{65.1} & {73.5} & \textbf{59.6}  \\
\bottomrule
\end{tabular}}\label{tab:general_benchmark_performance}
    \vspace{-3mm}

\end{table}

\subsection{Generalization to General Video Understanding Benchmarks}

To comprehensively evaluate our model's capabilities, 
we conduct extensive experiments on six challenging benchmarks spanning diverse aspects of video understanding: spatial reasoning (VSI-Bench~\cite{vsibench}), knowledge-intensive QA (VideoMMMU~\cite{videommmu} and MMVU~\cite{zhao2025mmvu}), and general video understanding (MVBench~\cite{li2024mvbench}, TempCompass~\cite{liu2024tempcompass}, and VideoMME~\cite{videomme}). For MMVU, we employ multiple-choice questions to ensure evaluation stability, while for VideoMME, we adopt the subtitle-free setting to focus on visual understanding.  

As shown in Tab.~\ref{tab:general_benchmark_performance}, our CARE-7B (SB-R1) achieves significant performance improvements over the base model across all benchmarks after training on SEED-Bench-R1. These consistent gains validate the quality of our benchmark's training data, the robustness of our methodology, and the comprehensiveness of our evaluation protocol.  

To further verify the effectiveness of our approach, we conduct additional experiments following Video-R1~\cite{video-r1}, training our model using GRPO-CARE with 16-frame video inputs on general-domain data (Video-R1-260k) for 1k RL steps and testing with 32-frame inputs. The comparative results from other methods shown in Tab.~\ref{tab:general_benchmark_performance} are taken from the Video-R1 paper. Notably, even when trained solely with RL, our model achieves competitive or superior performance compared to Video-R1-7B on most benchmarks. This is particularly remarkable given that Video-R1-7B benefits from explicit temporal order grounding constraints via GRPO rewards and supplementary supervised fine-tuning with additional data. Our model's ability to match or outperform this strong baseline with a more streamlined training pipeline underscores the efficiency of our method.  

Notably, our results suggest that improving reasoning-answer consistency can effectively encourage the model to align its responses with visual grounding results. This implicit approach demonstrates comparable efficacy to explicit visual perception constraints, presenting a promising alternative pathway for enhancing MLLM performance.

\section{Conclusion}
In this paper, we present SEED-Bench-R1, a comprehensive benchmark for evaluating post-training methods in MLLMs, and GRPO-CARE, a consistency-aware RL framework. SEED-Bench-R1 assesses model performance on tasks requiring balanced perception and reasoning through a novel three-level hierarchical generalization.
Our detailed analysis shows that while standard RL methods like GRPO improve answer accuracy, they often reduce reasoning coherence. GRPO-CARE addresses this by jointly optimizing correctness and logical consistency using likelihood calibration with a slow-evolving reference model, leading to better performance and interpretability.
We envision SEED-Bench-R1 and GRPO-CARE as valuable tools for advancing robust post-training methods, driving the development of more powerful MLLMs.

\clearpage

\medskip

\bibliography{neurips_2025}
\bibliographystyle{unsrt}

\end{document}